# Test Case Features as Hyper-heuristics for Inductive Programming

Edward McDaid[1][0000-0001-8684-0868] and Sarah McDaid[1][0000-0001-7643-6722]

[1] Zoea Ltd., 20-22 Wenlock Road, London N1 9GU, UK
edward.mcdaid@zoea.co.uk

**Abstract.** Instruction subsets are heuristics that can reduce the size of the inductive programming search space by tens of orders of magnitude. Comprising many overlapping subsets of different sizes, they serve as predictions of the instructions required to code a solution for any problem. Currently, this approach employs a single, large family of subsets meaning that some problems can search thousands of subsets before a solution is found. In this paper we introduce the use of test case type signatures as hyper-heuristics to select one of many, smaller families of instruction subsets. The type signature for any set of test cases maps directly to a single family and smaller families mean that fewer subsets need to be considered for most problems. Having many families also permits subsets to be reordered to better reflect their relative occurrence in human code – again reducing the search space size for many problems. Overall the new approach can further reduce the size of the inductive programming search space by between 1 and 3 orders of magnitude, depending on the type signature. Larger and more consistent reductions are possible through the use of more sophisticated type systems. The potential use of additional test case features as hyper-heuristics and some other possible future work is also briefly discussed.

**Keywords:** Inductive Programming, State Space Search, Hyper-heuristics.

## 1 Introduction

### 1.1 Background

The application of AI to software development is currently an active area [1]. Large language models [2] serve as the foundation for a growing number of systems that aim to support coding in a variety of ways [3]. While initial results appear promising, the long-term operational effectiveness of these tools remains to be seen. At the same time, there is a growing list of concerns about this technology including allegations of intellectual property misuse, bias, lack of transparency, veracity, security, stability, scalability, environmental and social impact [4, 5, 6]. In this context, the continued investigation of complementary and alternative approaches is prudent.

Inductive programming (IP) is a research field with the goal of turning a specification such as a set of test cases into code [7]. A major obstacle in achieving this is the need to search a huge space of candidate solutions. This is unavoidable - regardless of the approach employed - since the only way to determine the output of any non-trivial program is to execute it [8]. Thus, any system that automatically produces working code must incorporate a process that is equivalent to some form of search algorithm.

The size of the IP search space is largely determined by the branching factor, which in turn reflects the hundreds of instructions that make up high level programming languages. The term instruction denotes programming language operators, and core and standard library functions. A high branching factor causes the size of the search space to grow exponentially as program length increases. This means that even relatively small programs with just a few instructions can have enormous search spaces, making any form of exhaustive search infeasible. For decades, this has constrained the use of IP to the production of relatively small programs [9].

### 1.2 Zoea

Zoea is a knowledge-based inductive programming system that generates software automatically from a set of test cases [10]. It has been developed over the last few years with the key goal of significantly increasing the size of programs that can be created using IP. This has involved considerable ingenuity and innovation.

Zoea comprises a large number of knowledge sources, encapsulating various aspects of programming language and software development knowledge. These are integrated using a distributed blackboard architecture. Test cases are used ubiquitously as both the input specification and as the basis for all knowledge representation and reasoning.

Users can assemble larger programs from smaller ones using the associated Composable Inductive Programming paradigm and visual programming. Direct reasoning is used to avoid search where possible. Zoea also features a number of novel technologies that aggressively reduce the search space size.

The size of the IP search space is largely determined by the number of programming language instructions. Most modern high-level programming languages include over 200 instructions although individually the majority of programs use relatively few of these. Intuitively, if we could somehow predict the exact set of instructions that are required for a given problem then the corresponding search space would be exponentially smaller. Such an approach would still be useful even if it takes many attempts to predict the correct subset of instructions. This thinking motivated an earlier study of the instruction co-occurrence patterns that exist in human code [11].

### 1.3 Instruction Subsets

It has long been known that the frequency distribution of individual instructions in large samples of source code has a highly skewed, Zipfian distribution [12]. The previous study found that the distribution of instruction pairs is even more skewed [11]. That is, most possible pairs of instructions never or very rarely co-occur in human code. This result led to the development of the instruction subset approach.

Instruction subsets capture the patterns of instruction co-occurrence that exist within human code. They consist of many overlapping subsets of instructions that were derived from a large code sample. These were created by extracting the instruction subsets for program units (functions, subroutines and main blocks), which were then clustered to form families of subsets of specified sizes.

All Zoea knowledge sources support the provision of a custom instruction set. This means that any form of processing including search can be constrained by using each of the instruction subsets in turn.

Different families of instruction subsets were created, each with a given maximum size in terms of the number of instructions (e.g. 10, 20, ... 100). This allows search to begin by looking for solutions with 10 or fewer unique instructions. If this is unsuccessful then subsets of size 20 are tried, and so on. Approximately 90% of program units in human code use 10 or fewer unique instructions while only 2% have more than 20 [11]. Note that the number of unique instructions does not limit the number of instructions, so subset size does not constrain the size of generated programs.

The number of instruction subsets in families range from around 50 to 8000, depending on the maximum subset size. Within a given subset family each subset is used in turn to constrain search. Different subsets can also be processed in parallel using any number of computer cores.

Instruction subsets can shrink the search space by up to 40 orders of magnitude by reducing the branching factor from around 200 to 10 or 20 in most cases [11]. They do this because they encode the patterns of instruction co-occurrence used by human developers. In effect they leave out the many combinations of instructions that people never use. Also, the majority of problems can be solved within the first 5 subsets although there is a long tail for increasingly unusual combinations of instructions.

### 1.4 Instruction Digrams

Instruction subsets effectively capture tacit programming knowledge that describes what instructions can be used together. This work was continued in a second study that investigated how instructions can be combined to form solutions. This resulted in the development of instruction digrams [13].

Instruction digrams encode the patterns of instruction application present in human code. As such they represent another tacit form of coding knowledge. This information was also extracted from a large code sample. The result is a data flow model that records the frequency of every observed instruction-instruction application (or digram). Each digram basically corresponds to: *instruction A calls instruction B with a frequency F*. Put another way: *the output of instruction B is utilised as an input to instruction A with a frequency F*.

As is the case with instruction pair co-occurrence in instruction subsets, the vast majority of possible instruction digrams are never or rarely used by human developers. The frequency is recorded as a count and this can be thresholded for any value of *F* to identify a set of permissible instruction applications.

Instruction digrams are compiled into instruction subsets for deployment. For each subset family (of a given maximum size), this involves creating a variant of each instruction subset for each data flow depth in the search tree, from 1 up to some specified maximum depth. These additional instruction subset variants frequently have fewer instructions as depth increases - depending on the instructions present at the previous depth and the permissible digram rules. The removal of instructions at various depths has the effect of further reducing the size of the search space - in this case by up to five further orders of magnitude [13].

### 1.5 Possible Improvements

Each reduction in search space size means that larger programs can be generated with given resources. Our current approaches can be improved on in a number of ways. An obvious target is the number of subsets in subset families.

Instruction subsets impact the overall size of the search space in three ways. Firstly, the subset size largely determines the size of the subspace for each subset of that size. Secondly, the number of subsets determines the total size of the search space for a subset family. More aggressive clustering could reduce the number of subsets that need to be considered. Thirdly, the distance of frequently occurring subsets from the start of a family impacts the actual amount of the search space that must be explored.

### 1.6 Subset Reordering

We can attempt to reorder subsets to reflect their frequency distribution in human code. However, moving any one subset towards the start of a family will result in many other subsets being moved towards the end. With a single family of subsets, no ordering is possible that will address this problem for more than one subset at a time. Subset reordering would have more benefit if each subset family were to be decomposed in some way.

For any given problem there are often a large number of subsets that do not yield a solution. One obvious reason for this is that some subsets do not contain any instructions that produce the required data type as output. Clearly, a similar issue can occur with respect to program inputs however this can be less apparent. To address these issues we must consider the information available about data types relating to instructions, subsets and test cases.

### 1.7 Hyper-heuristics

Instruction subsets represent a form of heuristic [14]. A heuristic is a technique that is used to help to find any solution in a search space, where the cost of finding an optimal solution is prohibitive. In this sense, anything that reduces the size of the search space can be viewed as a heuristic. Instruction subsets have been shown to provide significant reductions in the size of the IP search space [11].

A hyper-heuristic is a technique that seeks to select, create or modify an optimum heuristic for a given problem [15]. Notably, hyper-heuristics work within a search space that is comprised of heuristics. In this sense, a family of instruction subsets represents a crude form of hyper-heuristic - as does the use of multiple families of instruction subsets of different maximum sizes. In both these cases the respective collections are simply iterated through until a solution is found - potentially using concurrency. A more powerful hyper-heuristic might use information available in the test cases to actively select the most relevant heuristics (instruction subsets).

## 2 Approach

### 2.1 Test Cases

Test case features can provide some clues about the required program. This can be useful even though the information is limited and uncertain. Each input and output data element in a test case is likely (but not certain) to play some role in the target program. While we can never know for sure, it is highly likely that most, if not all, inputs and outputs are in fact used. Degenerate cases do exist where programs ignore some or all inputs, or produce constant output but these tend to be rare, and separate mechanisms to handle such cases already exist within Zoea.

Test case inputs and outputs are respectively consumed by and produced by one or more instructions in a solution. As a result, the presence of a test case input element data type suggests that the solution code will require at least one instruction that takes that type as an input. Similarly, the data type of a test case output element signifies the need for at least one instruction that produces that kind of output. Thus, we can interpret test case inputs and outputs as providing a number of independent partial type signatures, for at least some of the instructions required for a solution.

Each instruction has a specification in terms of the input and output types that it supports. This allows us to infer possible instructions based on test case data types. This information would be of some use if we were selecting instructions from the complete instruction set. However, it is more useful in the context of instruction subsets that contain relatively small numbers of instructions.

### 2.2 Data Types

All test case data elements have an associated data type. In general terms this distinguishes between numbers, strings and structured data elements such as arrays. In some programming languages, data types can reflect both the nature of the data stored in a particular element (e.g. integer) and the size of the internal representation (e.g. short). There is no canonical set of data types and different languages use different classifications that can be rather general or very specific.

In this study we adopt a very general classification of data types based on those used in JSON [16]. We employ four data types namely number, string, array and hashmap (JSON object). Here, booleans and nulls are treated as strings. Adopting this approach makes the study simpler and also represents a lowest common denominator as far as types are concerned. The number of available data types determines the number of different combinations of data types that we can distinguish.

## 2.3 Type Signatures

A type signature is a compound element that describes the input and output data types of an instruction, test case or instruction subset. It is composed of a set of input types and a set of output types. The types number, string, array and hashmap are abbreviated as 'n', 's', 'a' and 'h' respectively. The input and output sets are separated using a colon or hyphen character. In order to simplify comparison the types are also sorted alphabetically within both the input and output sets. For example, the signature [a]:[s] represents the presence of an array input and a string output. The use of square brackets reflects the implementation of sets as lists. This signature can also be written 'a-s'.

Given the four data types (a, h, n and s), there are 15 ($2^4-1$) possible non-empty input and output type sets [a] to [a,h,n,s]. This gives 225 ($15^2$) possible type signatures ranging alphabetically from [a]:[a] to [a,h,n,s]:[a,h,n,s].

Different type signatures can be tested for equality. We say that a type signature *TS1* is subsumed by another type signature *TS2* if all of the input types in *TS1* are also present in *TS2* and all of the output types present in *TS1* are also present in *TS2*.

For test cases, the sets of data types of the input and output elements directly produce a type signature. With instruction subsets, the input type set is the set of data types associated with all of the input arguments of all of the subset instructions. Similarly the output type set is the set of output data types of all subset instructions.

Given the type signature for a test case and the type signature for an instruction subset, we can determine whether the instruction subset contains instructions that are capable of producing a solution - at least as far as the boundary data types are concerned. For any instruction subset, the role (if any) that each instruction might play in a candidate solution is unknown. In order for an instruction subset to be capable of producing a solution with a given output type, then at least one subset instruction must output that data type. Similarly, all of the input types required for the solution must be supported by some combination of the inputs of the subset instructions.

While this is not the same thing as producing a complete data flow graph, this approach enables us to quickly eliminate instruction subsets that are certainly incapable of producing a solution, due to the absence of one or more required input or output data types. It is still possible that some of the instruction subsets with compatible input-output types are not able to produce complete candidate solutions, although with a small number of data types this is less likely. There is scope for further work in this area with the possibility for some marginal search space reduction.

In this study the type signature does not consider the order or cardinality of input or output element types. Rather, we are only interested in the presence of a given type in the input or output part of a type signature. To this end each type occurs either once or not at all in the input and output sets.

## 2.4 Overview of Approach

Currently, a single family of instruction subsets is created for each maximum size. Each monolithic family contains all of the instruction subsets of a given size and they are used to solve every problem.

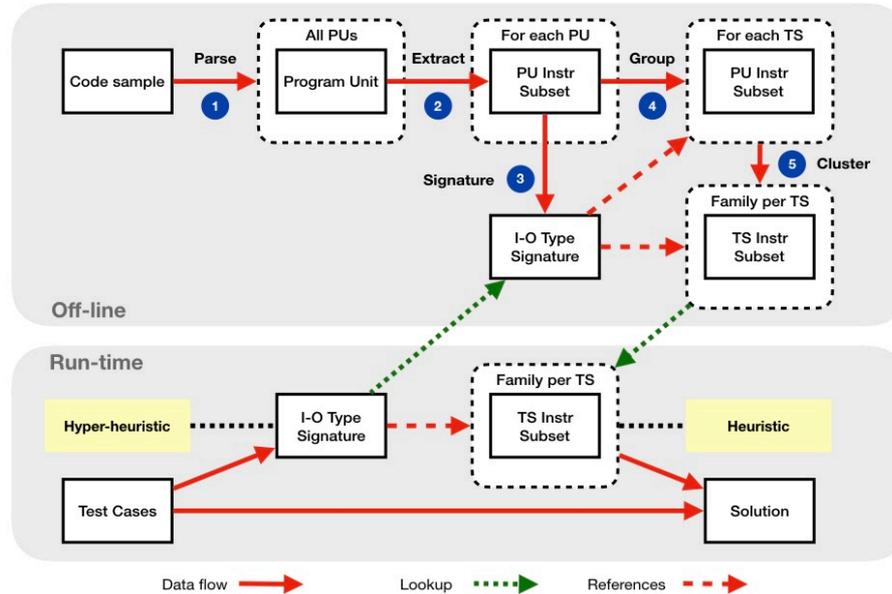

**Fig. 1.** Overview of the process used for type signature instruction subset family creation (upper pane) and run-time utilisation (lower pane).

In the new approach, multiple families are created for each maximum size - with each of these new families corresponding to a different input-output type signature. Each new family is constructed exclusively from program unit (PU) subsets with the same type signature. This ensures that at least some of the instructions in every instruction subset will be capable of consuming and producing the relevant input and output data types.

The alternative strategy of utilising PU subsets with a subsuming (see 2.3 above) type signature was deliberately rejected. This is because few PU subsets are excluded and the resulting instruction subset families are excessively large.

So now when we are presented with a problem - expressed as a set of test cases - we can create a type signature for the test cases and select which one of the new families of instruction subsets to use. This will ensure that only instruction subsets that are relevant to the problem type signature will be used. Figure 1 provides an overview of the approach.

We are also able to reorder the instruction subsets within each family. Some of the instruction subsets match more of the program units than others. This is because some of the program unit subsets are very common while others are rare. If we assume that the solutions to future problems have a similar instruction subset distribution to that of the code sample from which the instruction subsets were created, then reordering in this way should reduce the average number of subsets that need to be examined to produce solutions.

Within each family the instruction subsets are ordered in descending PU subset frequency. That is, the first subset in each family is the one with the largest number of corresponding program unit subsets, and the last subset has the lowest number of PU subsets.

### 2.5 Relative Non-productive Effort

In order to validate the approach we needed to demonstrate a reduction in the amount of work required in order to produce solutions. We did this by measuring and comparing the amount of work required to solve the same problems using both the old and new approaches. This allows us to quantify any reduction and determine whether or not the approach is successful, based on a target of achieving a reduction of at least one order of magnitude.

Now we already know from previous studies that the instruction subset approach works. For example, consider a problem that requires a given subset of instructions in order to produce a solution. We know that we will certainly be able to produce a solution if we encounter an instruction subset that contains all of the required instructions. This means that we do not need to undertake search and produce actual solutions in order to come up with a relative measure of the effort required. Instead we can simply count the number of instruction subsets that need to be examined before we find a subset that contains all of the required solution instructions.

For any given problem we can characterise the effort involved in finding a solution as either productive or non-productive. Searching one or more instruction subsets that do not contain the required instructions is non-productive effort. On the other hand searching an instruction subset that does result in a solution is productive effort. On this basis, counting the number of subsets at the start of a family that do not contain the target instructions (leading subsets) can provide a relative measure of the non-productive effort required. It is this figure that we would like to minimise.

This strategy of simply counting leading non-productive subsets also allows us use program unit instruction subsets as a stand-in for test cases. Since the instruction subsets are created from the PU subsets we can be certain that a solution exists for each case in at least one subset. The PU subsets are also numerous, diverse and highly representative of human code. Indeed, it would be difficult to find or manufacture a better set of test cases.

For each set of program unit subsets matching a given type signature we can measure the effort required using a single monolithic family of instruction subsets and compare this with the effort required using the corresponding type signature instruction subset family. While the instruction subsets correspond to search spaces with slightly different sizes, they are of similar size in terms of orders of magnitude. Also, any search involving an instruction subset that is unsuccessful will often involve an exhaustive search up to some predefined limit.

### 2.6 Scope

The current study was limited to subsets of size 10 only. This also means that only program unit subsets of size 10 or less were used, however this still accounts for over 90% of all program units. It is assumed that the approach would also be valid for larger subsets and program units although this would need to be confirmed in future work.

In any event, the findings would still be useful even if it turns out that the results do not apply to larger subsets.

Instruction digrams are excluded from the current study. This is because they produce a reduction in the search space size through a different mechanism. Also, the reduction that is provided by instruction digrams varies according to a number of factors and this would make any benefits arising from the use of type signatures more difficult to quantify. As a result the original instruction subset approach was used as the baseline against which any benefit arising from the use of type signatures are measured. There is every reason to believe that the benefits of digrams and type signatures would be cumulative although this would need to be verified in future work.

The code sample that was used is the same as that employed in the original instruction subsets study. This comprised the Python code contained in the largest 1000 repositories on GitHub at the time when the sample was retrieved. This represents about 15.75 million lines of code, corresponding to a diverse set of projects and domains.

Amplification is an optional step in subset creation that significantly improves the generality of the resultant subsets. This is accomplished before clustering by creating many additional artificial program unit subsets through the combination of smaller PU subsets. Amplification was enabled for the current study in order to ensure that the size of the new subset families was comparable with the baseline subset family.

### 2.7 Type Signature Instruction Subsets

For a given maximum subset size we created a family of instruction subsets for each input-output type signature. This was accomplished as follows. Each program unit was extracted in turn and the corresponding program unit instruction subset (PUIS) was produced in exactly the same manner as the original instruction subset study. The type signature for the PUIS was determined by identifying the set of all PUIS instruction input types and the set of all PUIS instruction output types. This gave us the tuple { PUIS type signature, PUIS } for each PUIS.

Once all program units had been processed we could consider each possible type signature in turn. For each type signature we assembled all of the corresponding PUIS and used them to produce a new family of type signature instruction subsets (TSIS) using the same process as the original instruction subset study. This gave us the tuple { type signature, TSIS family } for each type signature.

## 3 Results

### 3.1 Baseline

The baseline used was the monolithic instruction subset family of maximum size 10 that was produced in the original instruction subset study. This was compared against each TSIS family individually for each PUIS, for each PUIS type signature.

By definition each program unit instruction subset maps to a single type signature. On the other hand, some possible type signatures were not mapped to by any program units. This was because these combinations of instructions were not used by human developers in the large code sample that we used. Perhaps they might be found in a still larger sample. In any event, these type signatures were excluded from the study. In production such type signatures would be mapped to the baseline family but the probability that they would ever occur is presumably extremely low.

### 3.2  Measurement

We considered each PUIS type signature in turn. For each PUIS type signature we iterated through each PUIS with that signature. With each PUIS we first counted the number of (leading) subsets at the start of the baseline family that are not a superset of PUIS, up to and excluding the first subset that is a superset of PUIS (giving *R1*).

With the same PUIS we then considered the TSIS family (before reordering) with the corresponding type signature and similarly counted the number of leading subsets that are not a superset of PUIS (giving *R2*). Next we reordered the TSIS family in descending PU frequency. We then recounted the number of leading subsets in the reordered TSIS family that are not a superset of PUIS (giving *R3*).

*R1* is a relative measure of the redundant work required to find the (PUIS) solution using the baseline instruction subsets while *R2* is the redundant work required using the type signature instruction subsets without reordering and *R3* is the same measure with reordering. We recorded the *R1*, *R2* and *R3* results for each PUIS.

The *R1*, *R2* and *R3* results were then aggregated for each type signature to give *T1*, *T2* and *T3* respectively. *T1* is a relative measure of the total work required to find solutions for all program units with a given type signature using the baseline instruction subsets. *T2* is a similar measure using the corresponding TSIS family for the same type signature without reordering, while *T3* is the same measure with reordering. We recorded *T1*, *T2* and *T3* for each type signature.

Finally we divided *T1*, *T2* and *T3* for each type signature by the number of matching program unit subsets with the same type signature, giving *M1*, *M2* and *M3* respectively. *M1* is the mean relative work for each program unit using the baseline family for each type signature. *M2* is the mean relative work per program unit using the corresponding TSIS family but without reordering for each type signature, while *M3* represents the same measure with reordering. We recorded *M1*, *M2* and *M3* for each type signature.

### 3.3  Data

A detailed description of the data used in this study can be found in [11]. Figure 2 provides a perspective on the number of instruction subsets (Y1-axis) and matching program units (Y2-axis) for each type signature family (X-axis).

It can be seen that the number of instruction subsets in TSIS families varies considerably and that many have very few subsets. Also, the number of subsets in the largest TSIS family is around half that of the baseline family. It is also clear there is little correlation between the number of subsets and the number of matching program units. This is due to the fact that some PU signatures are very common even though they may correspond to a small number of different instruction subsets.

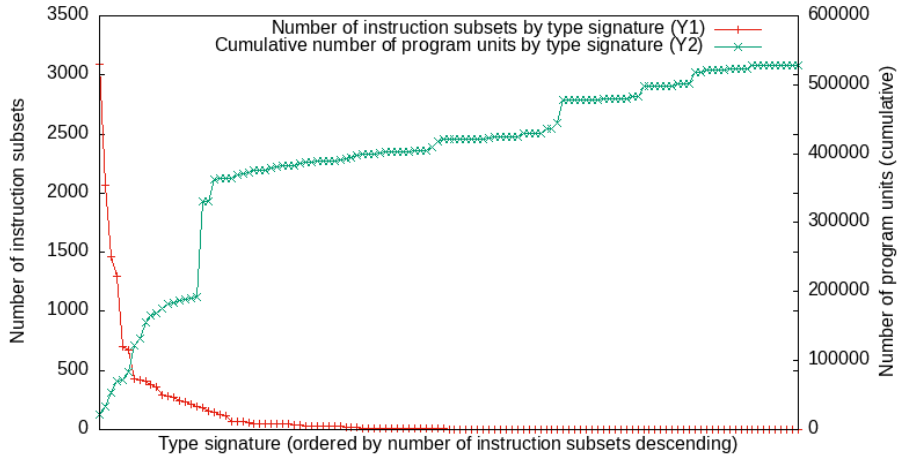

**Fig. 2.** Number of instruction subsets (red, Y1) and cumulative number of program units (green, Y2) for each type signature.

### 3.4 Search Space Reduction

Figure 3 shows the relative search space size reduction for each type signature instruction subset family. The red bars are the baseline (*M1*) for each signature. Data elements are arranged in descending order of baseline value. The green bars (*M2*) show the reductions corresponding to type signatures without reordering and the blue bars (*M3*) show the total reduction for each signature with type signatures and reordering. Please note that the Y-axis is log scale.

The first thing that is apparent is that there is considerable variation in the size of the baseline measure of over 2 orders of magnitude. Type signature families that correspond to many different program units often have larger numbers of instruction subsets. This is a reflection of the small number of types used in the signatures and can impact both simple type signatures (e.g. n-n) and those involving compound data structures.

There is also a wide range in the degree of both measures of reduction of up to 3 orders of magnitude. The smallest total reduction is around one order of magnitude. In addition, the reductions achieved through reordering are generally much smaller than those attributable to decomposition. Generally, the degree of both kinds of reduction increase as the baseline figure decreases. This suggests that higher baseline and lower reduction figures reflect insufficient family decomposition. This is to be expected with the simple type system that was used.

The results confirm that the approach of decomposing instruction subset families according to type signature hyper-heuristics does indeed work. They are also evidence that subset reordering can deliver a significant contribution to the total reduction.

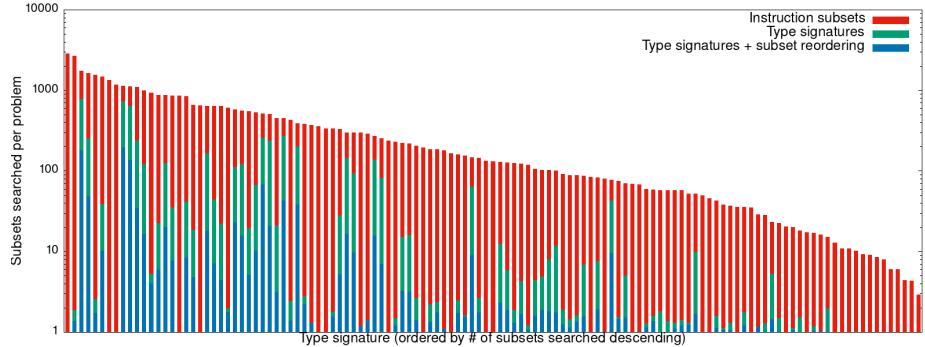

**Fig. 3.** Relative search space size reduction for each type signature instruction subset family, showing baseline (red), type signatures only (green) and type signatures plus reordering (blue).

## 4 Discussion and Future Work

The study demonstrates that the reductions in search space size obtained are the direct result of two processes. The first is the decomposition of larger subset families into many smaller families. This means that for any problem we need to consider many fewer subsets and that none of these are incapable of producing a solution. Furthermore, it significantly reduces the number of subsets required for long tail targets.

The second cause is subset reordering within families. It is easier to see why subsets that are used more frequently should be evaluated first. With a single monolithic family it is not possible to do this without also pushing everything else towards the back. Subset family decomposition is therefore a key enabler for subset reordering.

Larger and more consistent reductions could be obtained if we had a mechanism to map test cases to many more signatures. One way to achieve this would be through the use of a more sophisticated type system. We could also potentially use other test case features in addition to data types in order to produce signatures. Having more signatures means that we would also have more families and this in turn would have the potential to improve our ability to reorder subsets.

There are many test case features that could be used to generate signatures. For example, we might also consider the values of test case input and output elements. This could include lower and upper bounds on numeric values, the number of significant digits, the lengths of strings or the ratios of values across input and output elements. It would also be possible to test for the presence of numbers, dates or any other recognizable entities within strings. We could also observe whether and how the elements of an array are sorted, and so on. Indeed, a large number of options are available which could make it challenging to identify the most useful combinations of features for signature generation. Zoea already provides some automated feature engineering functionality that could be applied to this problem.

Ultimately, there is a limit to the amount of search space size reduction achievable through family decomposition and reordering. The number of subsets in the size 10 family is around 6000. This limits the reduction possible through fewer subsets to under 4 orders of magnitude. Beyond this further significant improvements could be achieved by altering the composition of subsets - as is done with instruction digrams. In any event, there is still work to do to achieve consistent levels of reduction.

The purpose of the current study was to demonstrate the feasibility of the approach and this has been accomplished. Subset production is currently largely an offline activity and integration of this work with the main codebase is not difficult. This chiefly involves adding the necessary support for instruction digrams, which is simply a matter of producing the subset variants for each TSIS family subset.

The current widespread interest in generative AI may lead some to question the need for this work or indeed it's relevance. Nevertheless, this study and the wider Zoea programme are attempting to achieve much the same goals and address similar problems while coming from a different direction. It is hard to see how some of the challenges faced by large language models in the context of code generation will be addressed without a deeper understanding of the knowledge used by software developers – both explicit and tacit. This work is helping to improve that understanding.

Some of the knowledge that Zoea uses is large, messy, and must be extracted in a more or less experimental manner. However, this does not prevent it from being formalised and made comprehensible. It could be argued that if Zoea and LLM based tools can achieve similar outcomes in generating code, then maybe in their own ways they both contain some of the same knowledge. In the case of Zoea, we know what that knowledge is.

As with the original instruction subset study upon which this current work is based, we believe the approach is completely ethical and does not conflict with any of the intellectual property rights of open source developers. The data extracted and used was limited to sets of instruction identifiers. These are not unique to any individual or project but rather occur frequently and are widely duplicated. This data therefore cannot represent anyone's intellectual property.

## 5     Conclusions

We have introduced the use of test case type signatures as hyper-heuristics to further reduce the size of the inductive programming search space. This work builds on the previously defined instruction subset approach that was also used as a baseline. The principle of operation involves determining the input and output data types present in a set of test cases and using this information to select one of many smaller instruction subset families. This improves over the previous approach by both significantly reducing the number of subsets that need to be considered, and by allowing subsets to be reordered within families to better reflect their frequency of use. A study was conducted to validate this technique. The study demonstrated a useful incremental reduction in search space size of between 1 and 3 orders of magnitude. When combined with our existing approaches this could reduce the search space size by up to 48 orders of magnitude. The results and some ideas for future work have been briefly discussed.

**Acknowledgements**

This work was supported entirely by Zoea Ltd. Zoea is a registered trademark of Zoea Ltd. Any other trademarks are the property of their respective owners.